\documentclass[conference]{IEEEtran}
\usepackage{cite}
\usepackage{amsmath,amssymb,amsfonts}
\usepackage{algorithmic}
\usepackage{graphicx}
\usepackage{textcomp}
\usepackage{makecell}
\usepackage{tabularray}
\usepackage{xcolor}
\usepackage{multicol}
\usepackage{multirow}
\usepackage{hyperref}
\IEEEoverridecommandlockouts
\IEEEpubid{\makebox[\columnwidth]{979-8-3503-0514-2/23/\$31.00~\copyright2023 IEEE \hfill} \hspace{\columnsep}\makebox[\columnwidth]{ }}
\def\BibTeX{{\rm B\kern-.05em{\sc i\kern-.025em b}\kern-.08em
    T\kern-.1667em\lower.7ex\hbox{E}\kern-.125emX}}
\begin{document}

\title{MLSD-GAN - Generating Strong High Quality Face Morphing Attacks using Latent Semantic Disentanglement  \\
}
\author{\IEEEauthorblockN{\textit{$Aravinda \:Reddy\:PN^1$,
$Raghavendra\: Ramachandra^2$,
$Krothapalli\:Sreenivasa\:Rao^3$ , 
$Pabitra\:Mitra^3$}}
\vspace{0.2cm}

\IEEEauthorblockA{ $^1$Advanced Technology Development Centre, IIT Kharagpur, Kharagpur, West Bengal, India}
\IEEEauthorblockA{$^2$Norwegian University of Science and Technology (NTNU),Norway}
\IEEEauthorblockA{$^3$Department of Computer Science and Engineering, IIT Kharagpur, Kharagpur, West Bengal, India}

}

\maketitle
\IEEEpubidadjcol

\begin{abstract}

Face-morphing attacks are a growing concern for biometric researchers, as they can be used to fool face recognition systems (FRS). These attacks can be generated at the image level (supervised) or representation level (unsupervised). Previous unsupervised morphing attacks have relied on generative adversarial networks (GANs). More recently, researchers have used linear interpolation of StyleGAN-encoded images to generate morphing attacks.
In this paper, we propose a new method for generating high-quality morphing attacks using StyleGAN disentanglement. Our approach, called MLSD-GAN, spherically interpolates the disentangled latents to produce realistic and diverse morphing attacks. We evaluate the vulnerability of MLSD-GAN on two deep-learning-based FRS techniques. The results show that MLSD-GAN poses a significant threat to FRS, as it can generate morphing attacks that are highly effective at fooling these systems.

\end{abstract}


\section{Introduction}
Despite its high accuracy, a Face Recognition System (FRS) is vulnerable to many attacks, one of which is a face morphing attack. Face recognition systems (FRS) are rapidly gaining prominence in our daily lives, finding applications in diverse domains ranging from smartphone unlocking to identity verification at border crossings. Enrollment in the FRS can be either supervised (new bank account opening on the bank premise) or unsupervised (new bank account opening from home). Enrollment in the unsupervised method is always prone to high risk, where the data subject might enroll a morphed, manipulated, printed, or electronically displayed image. These types of attacks must be averted at the enrollment level by using a strong attack detection mechanism. Therefore, in this work, we mainly focused on morphing attacks. Face-morphing attacks aim to generate a facial image that exhibits a strong perceptual and identifiable resemblance to the contributory data subjects used for morphing.  Therefore, enrolling the morph image in FRS can result in the successful verification of all contributory data subjects and thus provide multi-identity verification indicating security vulnerability.

Face morphing can be undertaken at two distinct levels: image-level and representation-level. Image-level morphing entails manipulating facial landmarks and blending texture attributes \cite{b14}. Conversely, representation-level morphing employs GAN architectures to interpolate face images and synthesize morphed attacks \cite{b15}. This representation-level approach was enabled by the development of GANs.

This study investigated the vulnerability of state-of-the-art FRS models to morphing attacks by creating new morphing techniques using StyleGAN. To achieve this, we used the disentangled latents of the StyleGAN. For the identity parts of the latent, a structure latent transfer direction is applied, and then latents are spherically interpolated and fed into the StyleGAN to synthesize the morphed face image, which we refer to as Morphing using Latent Semantic Disentanglement (MLSD-GAN). By assessing the susceptibility of various FRS models to MLSD-GAN-based attacks, our study demonstrated the remarkable ability of these attacks to circumvent FRS systems compared to conventional morphing techniques, highlighting their superior efficacy in compromising FRs security.

\vspace{-0.3cm}
\section{Related works}
\subsection{Image level techniques or Supervised methods}

In the past, crafting morphing attacks using facial landmark-based techniques entailed substantial supervision, alignment, and blending to produce the desired morphs. Common supervision-based techniques include free-form deformation (FFD) \cite{b23} and Delaunay triangulation-based morphing \cite{b24}. Owing to the artifacts caused by pixel/region-based morphing using supervised methods, many preprocessing techniques have been illustrated in \cite{b24}. Although commercial solutions like face fusion and FantaMorph \cite{b25} promote reduced manual involvement, they fail to meet the International Civil Aviation Organization's (ICAO) stringent face image quality standards, which are crucial for electronic Machine Readable Travel Documents (eMRTD) and biometric verification applications

\subsection{Representation level techniques or Unsupervised methods}

To overcome the cumbersome supervised morphed images, an automated approach for creating morph images using generative adversarial networks (GAN) was first proposed by \cite{b27} named MorGAN. MorGAN uses an encoder, decoder, and discriminator architecture to synthesize $64\times 64$ images. This method failed to meet ICAO image quality standards established for border control. Furthermore, \cite{b12} used StyleGAN and increased the spatial dimension of the synthesized images from $64\times64$ to $1024\times1024$ compared with MorGAN. Recently, \cite{b30} proposed diffusion autoencoder-based morphing, where two encoders, namely the semantic encoder and stochastic encoder, encode the semantic and stochastic parts of the latents for two images. Stochastic latents are linearly encoded, and semantic latents are spherically encoded. Although this method yielded fine-grained, high-quality morphed images, the sample generation process was very slow compared to that of GANs. Recently, \cite{b15} proposed an identity-driven GAN, where two contributory subjects were passed to the ResNet-50 model and the final output of ResNet-50 was reshaped into (18,512) to mimic that of the extended space of StyleGAN. However, the work presented in \cite{b15}  did not consider the following:1) the coarse, medium, and finer details of the latents of StyleGAN, and 2) injecting the averaged latents directly into StyleGAN to synthesize the image is a bottleneck.

Thus, motivated by the limitations of the existing method, we aimed to generate morphed images using StyleGAN latent semantic disentanglement, in which The StyleGAN \cite{b3} authors found that different style inputs correspond to different levels of detail. They classified these inputs into three groups: coarse, medium, and fine. Following this observation, \cite{b8} developed a feature pyramid network that generates three feature levels. Using this pyramid structure, \cite{b28} developed latent semantic disentanglement for swapping faces. We modified the method proposed in \cite{b28} to generate high-quality morphed images. Given an inverted latent, we split the latents into identity and attribute parts. From the decoupled latents, we derived the latent transfer direction for both contributory subjects. The modified latents were then spherically interpolated and post-processed to generate morphed images. Hence, the morphed images generated present a challenge for FRS detection. The key contributions of this study are as follows:
\begin{itemize}
    \item We present a new approach of generating morphed image through StyleGAN architecture latent semantic disentanglement.
    \item The proposed morph generation type is used to measure the attack success rate by verifying the deep learning-based FRS (Facial Recognition System) by performing the vulnerability test using a new dataset generated by using our morphed technique called MLSD-GAN.

    \item To illustrate the visual quality of the generated morphed image, perceptual quality metrics such as Peak Signal to Noise (PSNR) values are analysed.
\end{itemize}

The rest of the paper is organised as follows: Section \ref{sec:Proposed} discuss the proposed method, Section \ref{sec:Exp} presents the quantitative results of the proposed  and existing method. Finally, Section \ref{sec:conc} concludes the paper. 
\begin{figure*}
    \centering
    \includegraphics[width=1\textwidth]{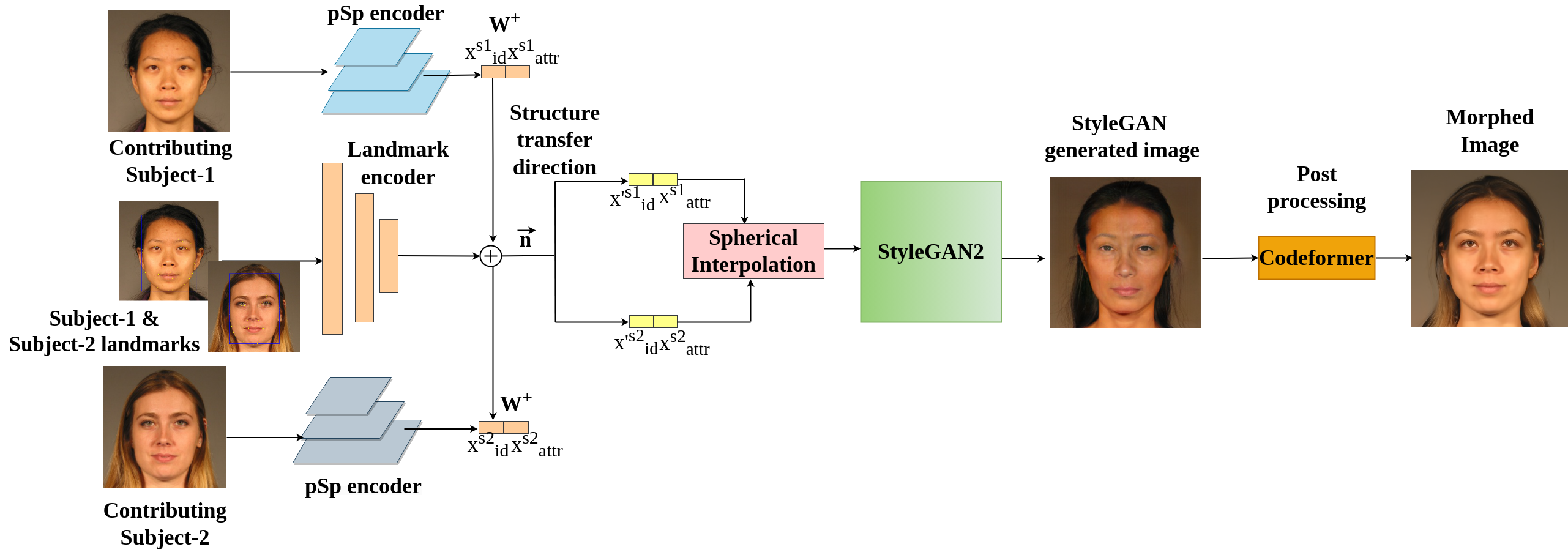}
    \caption{Block diagram of the proposed MLSD-GAN for generating high quality morphed face images}
    \label{fig:BD}
\end{figure*}

\begin{figure*}
    \centering
    \includegraphics[width=1\textwidth]{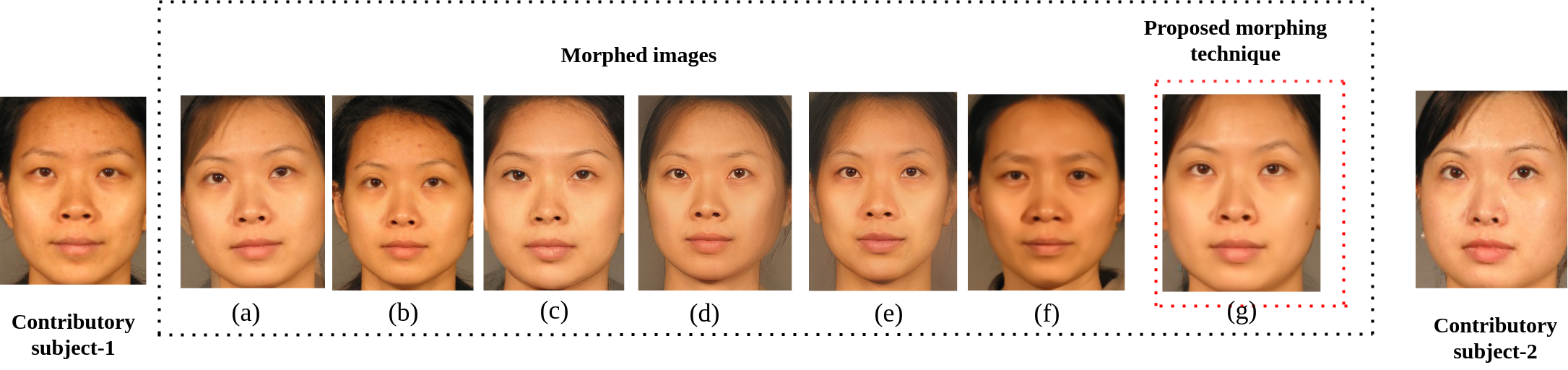}
    \caption{Qualitative results of the proposed MLSD-GAN with existing methods a) Landmarks-I b) Landmarks-II, c) StyleGAN, d) MIPGAN-1 e) MIPGAN-II, f) Morrdiff g) Proposed method}
    \label{fig:cmp_img}
\end{figure*}

\section{Method}
\label{sec:Proposed}
\subsection{Overview}

Figure \ref{fig:BD} shows the block diagram of the proposed morph generation technique. 
Given two high-resolution images, we aim to morph two images, subject-1 $x_{s1}$ and subject-2 $x_{s2}$ to obtain a morphed image by passing the subject-1 and subject-2 images to the pSp encoder \cite{b8}. The pSp framework was built on the representative power of the pre-trained StyleGAN generator and extended latent space. Given an input image we encode this image into $\mathbb{W}^+$ using 512-dimensional vector obtained from last layer of
the encoder network along with learning all 18 latent vectors.  However, learning all 18 style networks from such a network makes it difficult to learn all the finer details of the image, and the generated image quality is limited. The StyleGAN architecture distinguishes style inputs into three broad categories: coarse, medium, and fine inputs. Based on this observation, the pSp encoder was designed to generate three levels of feature maps. These feature maps were then used to extract styles using a convolutional mapper.  Therefore, the first K vectors of the latent code extracted from the shallow layers correspond to the identity features of the image, and the remaining latent vectors are extracted from deeper layers corresponding to the attributes of the image. Following the pSp encoder, we chose the first seven vectors for the identity part for both the subject-1 and subject-2 images, and the remaining 11 vectors were used for the attribute parts. So the latent codes for subject-1 image is denoted by $w_{s1}=(x_{id}^{s1},x_{attr}^{s1})$ and for subject-2 image $w_{s2}=(x_{id}^{s2},x_{attr}^{s2})$ where $x_{id}^{s1},x_{id}^{s2}$ are the identity parts and $x_{attr}^{s1},x_{attr}^{s2}$ are the attribute parts of the contributory images. To reconstruct the identity features of the morphed image that has the same identity as that of the subject-1 and subject-2 images, we must compute the identity transfer latent direction $\overrightarrow{n}$ which is derived from subject-1 and subject-2 landmark structures.

\begin{equation}
    \tilde{x}_{id}^s=x_{id}^{s1}+\overrightarrow{n}
\end{equation}

\begin{equation}
    \tilde{x}_{id}^t=x_{id}^{s2}+\overrightarrow{n}
\end{equation}

In order to determine the numerical coordinates of the facial landmarks $\overrightarrow{n}$, we employed a quantization technique known as randomized rounding, as introduced in \cite{b9}. This method effectively converts fractional parts of the coordinates into integers, enabling accurate and robust facial landmark localization through heat-map regression. In the traditional heatmap regression method, the ground-truth numerical coordinates of facial landmarks were quantized into heatmaps during the training stage. This process involved converting the continuous numerical values into discrete integer coordinates, which inevitably introduced quantization errors. To overcome these limitations, randomized rounding has been introduced as a more advanced quantization technique. Nevertheless, the quantization technique used in this method is either floor, round, or ceiling, which dumps the fractional parts of the numerical coordinates, thus making it difficult to reconstruct a heatmap. To derive the identity transfer direction, we pass the subject-1 and subject-2 landmarks to the landmark encoder. For both subject-1 and subject-2 we use 106 point facial landmarks rather than regular 58-point landmarks.

\subsection{Landmark Encoder}

The 106 point landmarks obtained from \cite{b9} are passed to the pSp encoder \cite{b8}. The pSp encoder inverts the image into an extended latent space $\mathbb{W}^+$ with feature pyramids at three levels: coarse, medium, and fine. The lower layers of the pSp encoder were used to produce the latent transfer direction $\overrightarrow{n}$.
\begin{equation}
    \overrightarrow{n}=E_{le}(l_{s1},l_{s2})
\end{equation}

\subsection{Spherical Interpolation}
\begin{figure}
    \centering
    \includegraphics[width=0.3\textwidth]{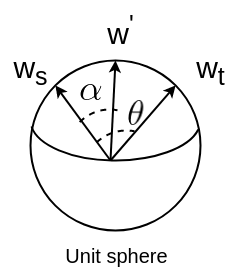}
    \caption{Spherical interpolation between subject-1 and subject-2 latents where $\alpha$ is the interpolation factor lies between 0 to 1 in our case $\alpha=0.5$ and $\theta$ is the angle subtended by the arc on the unit sphere.}
    \label{fig:slerp}
\end{figure}

A spherical interpolation operation was performed on the quaternions \cite{b10} to provide the smoothest interpolation between orientations. Spherical interpolation is the linear interpolation of the surface of a unit sphere. The idea behind spherical interpolation is simple: instead of moving over the shortest path (line, i.e., linear interpolation) from subject-1 latent ($w_{s1}$) to subject-2 latent ($w_{s2}$), we take the sphere arc path, as shown in Figure \ref{fig:slerp}. Spherical interpolation was performed on latents that captured the identity features of subject-1 and subject-2 images using formula \ref{eq:slerp}. After performing spherical interpolation on the latent-modified latents (18,512), they are fed into StyleGAN to obtain the morphed image and apply the post-processing technique named Codeformer to improve the quality of the morphed image.

\begin{equation}
    slerp(w_{s1},w_{s2},\alpha)=\frac{sin((1-\alpha)\theta)}{sin(\theta)}w_s+\frac{sin(\alpha\theta)}{sin\theta}w_t
    \label{eq:slerp}
\end{equation}

\subsection{Post-processing}
In \cite{b29}, a transformer-based code prediction network called CodeFormer was proposed to address the fundamental challenges of blind-face restoration. With the learned small discrete but expressive codebook space, we turned face restoration to code-token prediction, significantly reducing the degradation of face images. For this task, a transformer-based code prediction network was used to exploit the global composition of the low-quality images. The transformer module uses low-quality features as inputs and predicts the codebook sequence as codebook space. Benefiting from these designs, the codeformer shows great expressiveness and robustness against heavy degradation.

\section{Experiments and Results}
\label{sec:Exp}
This section delves into the experimental procedures, datasets, and quantitative outcomes associated with both the proposed and existing morphing techniques. The efficacy of face morphing generation is quantitatively assessed by gauging the Vulnerability of deep-learning-based FRS to morphed face images.

\subsection{Face Morph Dataset}
We employed face images from the FRGC-V2 face database \cite{b13} to generate morphed images using the proposed technique. For the purpose of this study, we meticulously curated a subset of 140 images from the FRGC dataset, giving preference to those that matched the high standards of passport image quality. The selected images comprised 47 female subjects and 93 male subjects. For each data subject, additional 7-21 images were captured for each subject, bringing the total number of images to 1270. All the samples underwent pre-processing to comply with ICAO standards, \footnote{ http://www.icao.
int/publications/Documents/9303 p9 cons en.pdf}, and morphing was performed in accordance with the guidelines outlined by \cite{b16}. For comparison, we used six different morph-generation technique blending techniques of landmarks called Landmarks-I \cite{b11}, landmarks with automatic color equalization \cite{b14} named Landmarks-II, StyleGAN \cite{b12} Identity-driven GAN named MIPGAN-1 and MIPGAN-2 \cite{b15} and the recent diffusion autoencoder-based method MorDiff \cite{b30}.

\subsection{Vulnerability Analysis}
This section delves into the vulnerability assessment of the proposed morphing technique. We evaluate the performance of our technique by attacking two open-source deep learning-based Face Recognition Systems (FRS). Open-source FRS systems include ArcFace \cite{b18} and MagFace \cite{b19}. The operational threshold is set at a False Matching Rate (FMR) of $1\%$ following the \cite{b16} guidelines.

Researchers have developed various metrics to assess the vulnerability of face recognition systems (FRSs) against morphing attacks. These metrics include the Mated Morph Presentation Match Rate (MMPMR), Fully Mated Morph Presentation Match Rate (FMMPMR) and Morph Attack Potential (MAP) \cite{b12}. MMPMR measures vulnerability based on independent attempts, while FMMPMR considers pairwise probe attempts from the same subject. MAP encompasses multiple FRS and pairwise probe attempts, but it presents vulnerability as a matrix rather than a single value. To address this limitation, \cite{b22} introduced the generalized morphing-attack potential (G-MAP), which provides a comprehensive and unified metric for assessing FRS vulnerability as a single number.


\subsection{Definition of G-MAP}

Let $\mathbb{T}$ denotes the set of paired images (also be denoted as number of probe attempts), let $\mathbb{F}$ denote the set of FRS, let $\mathbb{G}$ denote the morph attack generation algorithm, Let $\mathbb{M}_a$ denote face morphing image corresponding to $\mathbb{D}$, let $\tau_l$ similarity score from FRS (l), then G-MAP is defined as follows:

\begin{equation}
\begin{split}
    G-MAP=\frac{1}{\mathbb{|G|}}\sum_d^\mathbb{|G|}\frac{1}{\mathbb{|T|}}\frac{1}{\mathbb{|M}_a|}\; 
    \underset{l}{min}\; \\\sum_{i,j}^{\mathbb{|T|},\mathbb{|M}_a|}  [(S1_i^j>\tau_l) \wedge ....(Sk^j_i>\tau_l)] \times [(1-FTAR(i,l))]
    \label{eq:MAP}
\end{split}
\end{equation}

where $FTAR(i,l)$ is the failure to acquire the probe image sample in attempt i using FRS(l).


\begin{table*}[!ht]

\centering
\caption{Vulnerability analysis using G-MAP (Multiple Probe attempts) of ArcFace and MagFace @ FMR=1\% of FRS}
\label{tab:Gmap_vul}
\begin{tabular}{|ccccc|}
\hline
\multicolumn{5}{|c|}{\textbf{GMAP(\%) Multiple Probe Attempts}}  \\ \hline
\multicolumn{1}{|c|}{\multirow{7}{*}{\makecell{ArcFace\\
FRS=1\%}}} & \multicolumn{1}{c|}{\textbf{Method}} & \multicolumn{1}{c|}{\textbf{Male}} & \multicolumn{1}{c|}{\textbf{Female}} &\textbf{Combined}  \\ \cline{2-5} 
\multicolumn{1}{|c|}{}                  & \multicolumn{1}{c|}{Landmarks-I \cite{b11}} & \multicolumn{1}{c|}{93.06} & \multicolumn{1}{c|}{94.62} &95.62  \\ \cline{2-5} 
\multicolumn{1}{|c|}{}                  & \multicolumn{1}{c|}{Landmarks-II \cite{b14}} & \multicolumn{1}{c|}{86.26} & \multicolumn{1}{c|}{88.92} & 88.56 \\ \cline{2-5} 
\multicolumn{1}{|c|}{}                  & \multicolumn{1}{c|}{StyleGAN \cite{b12}} & \multicolumn{1}{c|}{66.67} & \multicolumn{1}{c|}{60.19} &65.47  \\ \cline{2-5} 
\multicolumn{1}{|c|}{}                  & \multicolumn{1}{c|}{MIPGAN-1 \cite{b15}} & \multicolumn{1}{c|}{90.16} & \multicolumn{1}{c|}{90.63} & 88.69 \\ \cline{2-5} 
\multicolumn{1}{|c|}{}                  & \multicolumn{1}{c|}{MIPGAN-2\cite{b15}} & \multicolumn{1}{c|}{91.26} & \multicolumn{1}{c|}{90.47} & 89.03 \\ \cline{2-5}
\multicolumn{1}{|c|}{}                  & \multicolumn{1}{c|}{MorrDiff\cite{b30}} & \multicolumn{1}{c|}{94.87} & \multicolumn{1}{c|}{93.68} & 93.25 \\ \cline{2-5}

\multicolumn{1}{|c|}{}                  & \multicolumn{1}{c|}{\textbf{Proposed}} & \multicolumn{1}{c|}{\textbf{93.03}} & \multicolumn{1}{c|}{\textbf{93.63}} &  \textbf{93.22}\\ \hline
\multicolumn{1}{|c|}{\multirow{6}{*}{\makecell{MagFace \\ FMR=1\%}}} & \multicolumn{1}{c|}{Landmarks-I \cite{b11}} & \multicolumn{1}{c|}{94.09} & \multicolumn{1}{c|}{95.60} &94.36  \\ \cline{2-5} 
\multicolumn{1}{|c|}{}                  & \multicolumn{1}{c|}{Landmarks-II\cite{b14}} & \multicolumn{1}{c|}{87.62} & \multicolumn{1}{c|}{89.17} & 88.26 \\ \cline{2-5} 
\multicolumn{1}{|c|}{}                  & \multicolumn{1}{c|}{StyleGAN \cite{b12}} & \multicolumn{1}{c|}{76.67} & \multicolumn{1}{c|}{72.67} & 68.67 \\ \cline{2-5} 
\multicolumn{1}{|c|}{}                  & \multicolumn{1}{c|}{MIPGAN-1 \cite{b15}} & \multicolumn{1}{c|}{85.36} & \multicolumn{1}{c|}{86.46} &84.19  \\ \cline{2-5} 
\multicolumn{1}{|c|}{}                  & \multicolumn{1}{c|}{MIPGAN-2 \cite{b15}} & \multicolumn{1}{c|}{89.44} & \multicolumn{1}{c|}{87.62} & 88.09 \\ \cline{2-5} 
\multicolumn{1}{|c|}{}                  & \multicolumn{1}{c|}{MorrDiff\cite{b30}} & \multicolumn{1}{c|}{91.87} & \multicolumn{1}{c|}{92.68} & 90.25 \\ \cline{2-5}
\multicolumn{1}{|c|}{}                  & \multicolumn{1}{c|}{\textbf{Proposed}} & \multicolumn{1}{c|}{\textbf{91.09}} & \multicolumn{1}{c|}{\textbf{92.07}} &\textbf{92.03}  \\ \hline
\end{tabular}
\end{table*}

\begin{table*}[]
\centering
\caption{Vulnerability analysis using G-MAP metric (Multiple FRS and multiple probe attempts)}
\label{tab:G-MAP_mfrs}
\begin{tabular}{|cccc|}
\hline
\multicolumn{4}{|c|}{G-MAP(\%) (Multiple FRS and multiple probe attempts}                                                    \\ \hline
\multicolumn{1}{|c|}{Method} & \multicolumn{1}{c|}{Male} & \multicolumn{1}{c|}{Female} & Combined \\ \hline
\multicolumn{1}{|c|}{Landmarks-I \cite{b11}} & \multicolumn{1}{c|}{93.06} & \multicolumn{1}{c|}{94.62} &94.36  \\ \hline
\multicolumn{1}{|c|}{Landmarks-II \cite{b14}} & \multicolumn{1}{c|}{86.26} & \multicolumn{1}{c|}{88.92} &88.26  \\ \hline
\multicolumn{1}{|c|}{StyleGAN \cite{b12}} & \multicolumn{1}{c|}{66.67} & \multicolumn{1}{c|}{60.19} &65.47  \\ \hline
\multicolumn{1}{|c|}{MIPGAN-1 \cite{b15}} & \multicolumn{1}{c|}{85.36} & \multicolumn{1}{c|}{86.46} &84.19  \\ \hline
\multicolumn{1}{|c|}{MIPGAN-2 \cite{b15}} & \multicolumn{1}{c|}{89.44} & \multicolumn{1}{c|}{87.62} &88.09  \\ \hline
\multicolumn{1}{|c|}{MorDiff \cite{b30}} & \multicolumn{1}{c|}{91.87} & \multicolumn{1}{c|}{92.68} &90.25  \\ \hline
\multicolumn{1}{|c|}{\textbf{Proposed}} & \multicolumn{1}{c|}{\textbf{91.04}} & \multicolumn{1}{c|}{\textbf{92.07}} &\textbf{92.03}  \\ \hline
\end{tabular}
\end{table*}

Different parameters can be used to compute G-MAP, such as multiple attempts, multiple FRS, and morph attack generation types. G-MAP with multiple probe attempts was calculated using \ref{eq:MAP} by setting D=1 and F=1. Thus, G-MAP with multiple probe attempts is identical to FMMPMR \cite{b12} when FTAR=0. Furthermore, G-MAP with multiple FRS and multiple attempts is computed by taking the logical AND operation between different FRS using D=1. Thus, using G-MAP, a single vulnerability value was obtained.

\subsection{Quantitative Evaluation of G-MAP}
Since G-MAP is influenced by probe attempts, FRSs, and morphing types, we can analyze the quantitative results pertaining to (a) how probe attempts impact FRS vulnerability in relation to the type of morphing attack used, (b) the overall vulnerability of multiple FRSs with multiple probe attempts, regardless of the morphing attack type, and (c) G-MAP as a function of attempts, multiple FRSs, and various morph attack types, including FTAR. 
\subsubsection{G-MAP with multiple probe attempts}
\begin{figure}[!h]
    \centering
    \includegraphics[width=0.55\textwidth]{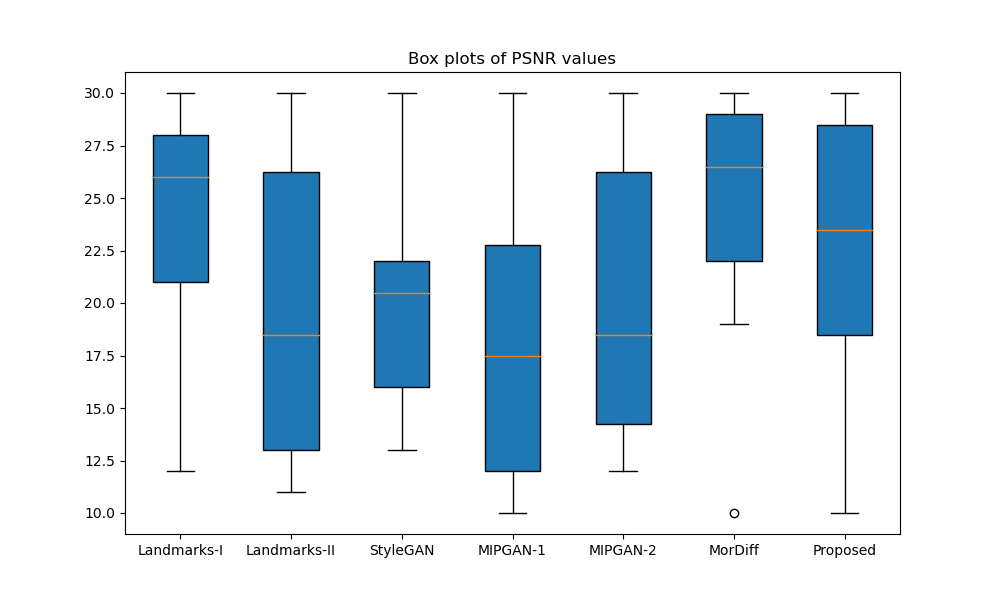}
    \caption{Box plots of PSNR values computed from different
face morph generation methods}
    \label{fig:psnr}
\end{figure}

The obtained success rate or vulnerability of FRS is provided in the Table \ref{tab:Gmap_vul}. Vulnerability analysis was performed using six different morph-generation methods: Landmarks-I, Landmarks-II, StyleGAN, MIPGAN-1, MIPGAN-2, MorDiff \cite{b30} and the proposed method. Extrapolating from the analysis revelations, these observations were inferred.
\begin{itemize}
    \item Amongst the two facial recognition systems, ArcFace \cite{b18} stands out as the most vulnerable to morphing attacks, exhibiting susceptibility to all morphing techniques.
    \item The proposed MLSD-GAN shows highest vulnerability with ArcFace FRS compared to StyleGAN \cite{b12} and MIPGAN-1 and MIPGAN-2.
    \item The proposed MLSD-GAN shows slightly higher vulnerability when compared to Landmarks-II.
    \item The MorDiff morph generation technique shows highest vulnerablity when compared to all other methods when FRS is ArcFace but using MagFace MorDiff morph generation technique vulnerability lags behind the proposed method slightly.
    \item The proposed MLSD-GAN with MagFace FRS shows highest vulnerability compared to StyleGAN and slighly higher vulnerability with MIPGAN-1 and MIPGAN-2.
    \item Overall the proposed MLSD-GAN shows higher vulnerability when compared to all other morphing methods when FMR=1\%.
    \item Figure \ref{fig:psnr} indicates the box plots of the PSNR values. These box plots indicate that the perceptual quality of the proposed method is superior to that of StyleGAN, MIPGAN-1, and MIPGAN-2.
\end{itemize}

\subsubsection{G-MAP with multiple FRS and multiple probe attempts}
In this type of analysis, we perform a logical operation between the multiple FRS systems, as shown in Table \ref{tab:G-MAP_mfrs}. The following observations are made from Table \ref{tab:G-MAP_mfrs}.

\begin{itemize}
    \item The vulnerability of the FRS is presented as single number with respect to multiple FRS. 
    \item Landmarks-I based morphing attack has the highest vulnerability when compared to all other morphing attacks.
    \item The proposed MLSD-GAN based morphing attack indicate highest vulnerability when compared to StyleGAN, MIPGAN-1, MIPGAN-2 morphing attack.
\end{itemize}
Figure \ref{fig:cmp_img} presents the qualitative results of the proposed method.

\section{Conclusion}
\label{sec:conc}
In this work, we propose a new morph generation technique called MLSD-GAN, which addresses the limitations of the previous morph generation technique. The proposed approach for deriving strong morphs was created using the disentanglement property of StyleGAN with a clear distinction with respect to the identity and attribute features of the image that were encoded and spherically interpolated to generate realistic images that threaten the existing state-of-the-art FRS. Vulnerability analysis was performed using two deep learning-based FRS, and the empirical results show that the existing state-of-the-art FRS are highly vulnerable, especially when compared to StyleGAN and variants of MIPGAN at @FMR=1\%. Future research will look into the generalization benefits of including MLSD-GAN-based morphs in training Morph Attack Detection solutions.




\vspace{12pt}
\color{red}

\end{document}